\documentclass{article}

\usepackage{times}
\usepackage{ijcai16}
\usepackage{url}
\usepackage{graphicx}
\usepackage{amsmath}
\usepackage[ruled,noline,slide,noend]{algorithm2e}
\usepackage{tabulary}
\usepackage{ragged2e}

\title{Multi-Document Abstractive Summarization \\Using ILP based Multi-Sentence Compression}
\author
{Siddhartha Banerjee \\ {The Pennsylvania State University}\\{University Park, PA, USA}\\ {sub253@ist.psu.edu}
\And
Prasenjit Mitra\\{QCRI} \\ {Doha, Qatar}\\ {pmitra@qf.org.qa}
\And
Kazunari Sugiyama\\{National University of Singapore}\\ {Singapore}\\ {sugiyama@comp.nus.edu.sg}}

\begin{document}
\maketitle
\begin{abstract}
Abstractive summarization is an ideal form of summarization since it can
synthesize information from multiple documents to create concise informative summaries.
In this work, we aim at developing an abstractive summarizer. 
First, our proposed approach identifies the most important document in the multi-document set. The sentences in the most important document are aligned to sentences in other documents to generate clusters of similar sentences. Second, we generate $K$-shortest paths from the sentences in each cluster using a word-graph structure. Finally, we select sentences from the set of shortest paths generated from all the clusters employing a novel integer linear programming (ILP) model with the objective of maximizing information content and readability of the final summary. Our ILP model represents the shortest paths
as binary variables and considers the length of the path, information score and linguistic quality score in the objective function. Experimental results on the DUC 2004 and 2005 multi-document summarization datasets show that our proposed approach outperforms all the baselines and state-of-the-art extractive summarizers as measured by the ROUGE scores. Our method also outperforms a recent abstractive summarization technique. In manual evaluation, our approach also achieves promising results on informativeness and readability.   
\end{abstract} 

\section{Introduction}\label{Sec:Intro}
Abstractive summarization has gained popularity due to its ability of generating new sentences to convey the important information from text documents. 
An abstractive summarizer should present the summarized information in a coherent form that is easily readable and grammatically correct. Readability or linguistic quality is an important indicator of the quality of a summary.
Several text-to-text (T2T) generation techniques that aim to generate novel text from textual input have been developed~\cite{knight2002summarization,zajic2007multi,Barzilay05sentencefusion}. However, to the best of our knowledge, none of the above methods explicitly model the role of linguistic quality and only aim at maximizing information content of the summaries. In this work, we address readability by assigning a log probability score from a language model as an indicator of linguistic quality. More specifically, we build a novel optimization model for summarization that jointly maximizes information content and readability. 
  
Extractive summarizers~\cite{mani1999advances} often lose a lot of information from the input as they only ``extract'' a few important sentences from the documents to create the final summary. 
%Genest and Lapalme~\shortcite{genest2010text} pointed out the empirical limit to extractive summarization. 
We prevent information loss by aggregating information from multiple sentences. We generate clusters of similar sentences from a collection of documents.
Multi-sentence compression (MSC)~\cite{filippova2010multi} can be used to fuse information from sentences in a cluster. However, MSC might generate sentences that convey similar information from two different clusters. By contrast, our Integer Linear Programming (ILP) based approach prevents redundant information from being included in the summary using a inter-sentence redundancy constraint. Consequently, our experiments reveal that our method generates more informative and readable summaries than MSC.
   
Our proposed approach to abstractive summarization consists of the following two steps: 
(1) Aligning similar sentences from multiple-documents and  
(2) Generating the most informative and linguistically well-formed sentence from each cluster, and then appending them together.
In multi-document summarization, all documents are not equally important; some documents contain more information on the main topics in the document set. % The documents close to the main topics of the document set are usually more important.
Our first step estimates the importance of a document in the whole dataset using \textit{LexRank}~\cite{erkan2004lexrank}, \textit{Pairwise Cosine Similarity} and \textit{Overall Document Collection Similarity}.
Each sentence from the most important document are
initialized into separate clusters.
Thereafter, each sentence from the other documents are assigned to the cluster that has the highest similarity with the sentence. %In a multi-document scenario, redundancy is an important component that helps decide content that is summary-worthy. 
In the generation step, we first generate a word-graph structure from the sentences in each cluster and construct $K$ shortest paths from the graph between the start and end nodes. 
We formulate a novel integer linear programming (ILP) problem that maximizes the information content and linguistic quality of the generated summary. 
%Kaz-IJCAI-1: The active mode is better. 
%Each shortest path in $K$ is represented as a binary variable in our ILP problem. 
Our ILP problem represents each of the $K$ shortest paths as a binary variable. 
The coefficients of each variable in the objective function is obtained by combining the information score of the path and the linguistic quality score. We introduce several constraints into our ILP model. We ensure that only one sentence is generated from each cluster. Second, we avoid redundant sentences that carry the same or similar information from different clusters. The solution to the optimization problem decides the paths that would be included in the final abstractive summary.

On the DUC2004 and DUC2005 datasets, we demonstrate the effectiveness of our proposed method. Our proposed method outperforms not only some popular baselines 
but also the state-of-the-art extractive summarization systems. 
ROUGE scores~\cite{lin2004rouge} obtained by our system outperforms the best extractive summarizer on both the datasets. 
Our method also outperforms an abstractive summarizer based on multi-sentence compression~\cite{filippova2010multi} when measured by ROUGE-2, ROUGE-L and ROUGE-SU4 scores. % [0.11992 vs 0.10612 -- ROUGE-2].
Further, manual evaluation by human judges shows that our technique produces summaries with acceptable linguistic quality and high informativeness.  

\section{Related Work}\label{Sec:RelWork}
Several researchers have developed abstractive summarizers. 
Genest and Lapalme~\shortcite{genest2010text} used natural-language-generation (NLG) systems. 
However, NLG requires a lot of manual effort in terms of defining schemas as well as using deeper natural language analysis. 
Wang and Cardie~\shortcite{wang2013domain} and Oya \textit{et al.}~\shortcite{templateSumm2014} induced templates from the training set in their meeting summarization tasks. 
Such induction of templates, however, is not very effective in news summarization because of the variability in topics. 
Unlike these methods, our method does not induce any templates but generates summaries in an unsupervised manner by combining information from several sentences on the same topic. 
Berg-Kirkpatrick \textit{et al.~}~\shortcite{Berg-Kirkpatrick:2011:JLE:2002472.2002534} used an ILP formulation that jointly extracts and compresses sentences to generate summaries. 
However, their method is supervised and requires significant manual effort to define features for subtree deletions, which is required to compress sentences.
Graph-based techniques have also been very popular in summarization.
Ganesan \textit{et al.}~\shortcite{ganesan2010opinosis} employed a graph-based approach 
to generate concise abstractive summaries from highly redundant opinions. 
Compared with their opinionated texts such as product reviews, the target documents in multi-document summarization do not contain such high level of redundancy.

More recently, Mehdad~\textit{et al.}~\shortcite{mehdad2013abstractive} proposed a supervised approach for meeting summarization, 
in which they generate an entailment graph of sentences. The nodes in the graph are the linked sentences and edges are the entailment relations between
nodes; such relations help to identify non-redundant and informative sentences. Their fusion approach used MSC~\cite{filippova2010multi}, which generates an informative sentence by combining several sentences in a word-graph structure.
However, Filippova's method produces low linguistic quality as the ranking of generated sentences is based on edge weights calculated only using word collocations. 
By contrast, our method selects sentences by jointly maximizing informativeness and readability and generates informative, well-formed and readable summaries.  

\section{Proposed Approach}\label{Sec:Method}
\begin{figure}[t]
  \centering
    \fbox{\includegraphics[width=0.35\textwidth, height=2.6in, keepaspectratio=false]{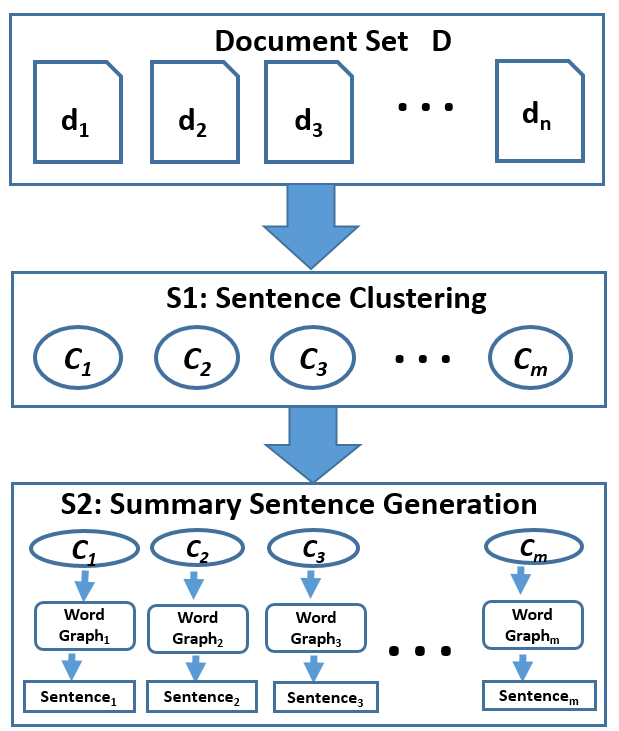}}
\caption{\small{Overview of our abstractive summarization approach.}}\label{approach-general}
\end{figure}

Figure~\ref{approach-general} shows our proposed abstractive summarization approach, which consists of the following two steps: 
\begin{itemize}
 \item[S1:] \textit{Sentence clustering},
 \item[S2:] \textit{Summary sentence generation}. 
\end{itemize}

Given a document set $\boldsymbol{D}$, that consists of $n$ documents ($d_{1}$, $d_{2}$, $d_{3}$, ..., $d_{n}$), our approach first generates $m$ clusters ($C_{1}$, $C_{2}$, $C_{3}$, ..., $C_{m}$) of similar sentences, and then use the individual clusters to create word-graphs. A maximum of one novel sentence is generated from each word-graph with the goal of maximizing information content and linguistic quality of the entire summary. 
The sentence clustering step (S1) has two important components: the first (S1-1) identifies the most important document $D_{imp}$ in $\boldsymbol{D}$ before the final cluster generation step (S1-2) that generates clusters of similar sentences. % (see 
%The summary sentence generation step (S2) generates a sentence from each of the clusters. 
We experiment with several techniques to identify $D_{imp}$, and then align sentences from other documents to the sentences in $D_{imp}$. It proves to be a simple, yet effective technique for generating clusters containing similar information. Our approach is inspired by the findings of Wan~\shortcite{wan2008exploration} that showed how the incorporation of document impact can improve the performance of summarization.
In (S2), we create a directed word-graph structure from the sentences in each cluster. %The construction of the word-graph was adapted from Fillipova's MSC approach~\cite{filippova2010multi} where 
In the word-graph, the nodes represent the words and the edges define the adjacency relations in the sentences. From the word-graph, multiple paths between the start and the end nodes can be extracted. We extract $K$ shortest paths from each cluster, and finally retain the paths that maximize information content and linguistic quality using an ILP based approach. We impose constraints on the maximum number of sentences that are generated from each cluster and also impose constraints to avoid redundancies such that similar information from different clusters are not included in the summary. Information content is measured using TextRank~\cite{mihalcea2004textrank}, which scores sentences based on the presence of keywords. We measure linguistic quality using a 3-gram language model that assigns confidence values to sequences of words in the sentences.
In this section, we describe both steps -- S1 and S2.

%Kaz-IJCAI-1: ``S1'' denotes a kind of headline of the section. So we can skip ``3.1'' here to remove duplicate section ID. 
%I think ``S1: ...'' looks better than ``3.1 S1: ...''  
\subsection*{S1: Sentence Clustering}
\label{sec:approach}
We initialize clusters of sentences using each sentence from the most important document, $D_{imp}$, in a document set $\boldsymbol{D}$. Our intuition behind this approach is that $D_{imp}$ consists of the most important content relevant across all the documents in $\boldsymbol{D}$. In other words, the document that is most close to the central content of the collection is the most informative.

\subsubsection{(S1-1) Document Importance}\label{subsec:DocImportance}
We propose several techniques to identify $D_{imp}$.

\noindent{\textbf{LexRank (${MD^{Imp}_{LexRank}}$):}}
%This method constructs a weighted graph between the documents and ranks the documents. 
LexRank~\cite{erkan2004lexrank} constructs %a modification of PageRank~\cite{page1999pagerank}, applied to 
a graph of sentences where the edge weights are obtained by the inter-sentence cosine similarities.
While the original LexRank constructs a graph of sentences, we construct a graph of documents to compute document importance. 
%The continuous version of LexRank relies on the strength of the similar links.
Equation (\ref{lpr-equation2}) shows how LexRank scores are computed using weighted links between the nodes (documents). 
%In our approach, we use cosine similarity between TF-IDF feature vector to obtain pairwise-similarity between all the sentences in our dataset. 
This equation measures the salience of a node in the graph, which is the importance of the document in the entire document collection.
Let $p(u)$ be the centrality of node $u$. LexRank is then defined as follows:
\begin{equation}
\label{lpr-equation2}
\resizebox{.46\textwidth}{!}
{$p(u) =  \frac{d}{N}+{(1-d)}\sum_{v\in adj[u]}\frac{\textrm{idf-modified-cosine}(u,v)}{\sum_{z\in adj[v]}\textrm{idf-modified-cosine}(z,v)}p(v),$} 
\end{equation}
where $adj[u]$ and $N$ are the set of nodes that are adjacent to $u$ and the total number of nodes in the graph, respectively. 
The damping factor $d$ is usually set to 0.85, and we set $d$ to this value in our implementation. $D_{imp}$ is determined as the document that has the highest LexRank score in $\boldsymbol{D}$ once the above equation converges.   
%We use cosine similarity as the link weight. 

\noindent{\textbf{Pairwise Cosine Similarity (${MD^{Imp}_{CosSim}}$):}} 
This method computes the average cosine similarity between the target document $d_{i}$ and the other documents in the dataset.  
The average similarity is calculated using the following formula:
\begin{displaymath} 
AveCosSim(d_{i})=\frac{\sum\limits_{d_{i},d_{j}\in D}{CosSim(d_{i}, d_{j})}}{|\boldsymbol{D}|-1}\quad (i\neq j), 
\end{displaymath} 
\normalsize
%In the above equation 
where $|\boldsymbol{D}|$ denotes the number of documents in the document set $\boldsymbol{D}$. 

\noindent{\textbf{Overall document collection similarity (${MD^{Imp}_{DocSetSim}}$):}}
This method computes the cosine similarity between the target document and the whole document set. We create the whole document set by concatenating text from all the documents 
in $\boldsymbol{D}$. This method is defined as follows:
\begin{displaymath} 
DocSetSim(d_{i}) = CosSim(d_{i}, \boldsymbol{D}). 
\end{displaymath}

\normalsize

\noindent In $MD^{Imp}_{LexRank}$, $MD^{Imp}_{CosSim}$, and $MD^{Imp}_{DocSetSim}$ mentioned above, 
we select the document $D_{imp}$ with the highest score as the most important one in the dataset $\boldsymbol{D}$. Next, we generate the clusters by aligning sentences and re-ordering them based on original positions of the sentences in the documents. 

\subsubsection{(S1-2) Cluster Generation}
%The assignment is based on the cosine similarity scores between any pair of sentences. 
The sentences from each of the other documents ($d_{i} \neq D_{imp}$) in $\boldsymbol{D}$ are assigned to the clusters one-by-one based on cosine similarity measure. 
%Pairwise cosine similarity of each sentence d\textsubscript{i}s\textsubscript{k} in d\textsubscript{i} is computed against all the sentences in \textit{Doc\textsubscript{imp}}. 
Our approach computes pairwise cosine similarity of each sentence in $d_{i}$ to all the sentences in $D_{imp}$.
%Kaz-IJCAI-1: Maybe, you can write this sentence more simple as follows:
For example, a sentence in $d_{i}$, $s_{p}^{d_{i}}$ has the highest similarity with $s_{j}^{D_{imp}}$, a sentence in $D_{imp}$.
Then, we assign $s_{p}^{d_{i}}$ to cluster $C_{j}$, in which $s_{j}^{D_{imp}}$ was initially assigned. %to in the initialization step. 
Some sentences in $d_{i}$ might not be similar to any of the sentences in $D_{imp}$. Hence, we only align sentences %that have a minimum similarity of $0.5$. 
when the similarity $> 0.5$.
%Kaz-IJCAI-1: Please check ``0.5$|D|$'' Is this  ``0.5$\times |D|$''?
Further, we only retain clusters that have at least $|D|/2$ sentences, assuming that a content is relevant only if it exists in half of the documents in $\boldsymbol{D}$.

\noindent{\textbf{Cluster ordering:}} 
We implement two \textit{cluster ordering} techniques that reorder clusters based on the original position of the sentences in the documents. 
\begin{enumerate}
% \item \textbf{Majority ordering (CO-1):} 
 \item \textit{Majority ordering (MO):}
%Majority ordering is one of the traditional ways of sentence ordering. In this work, we extend it to clusters. %This ordering technique assumes that clusters belong to different topics, and the ordering of clusters can be determined by the occurring sequence of themes in the source documents. 
Given two clusters, $C_{i}$ and $C_{j}$, the set of common documents from which the sentences are assigned to the two clusters are identified. 
If $C_{i}$ and $C_{j}$ have sentences $s_{p}^{d_{i}}$ and $s_{q}^{d_{i}}$ ($p < q$), respectively, where $d_{i}$ is the common document, then $C_{i}$ precedes $C_{j}$. %Ties are broken by a 
The final order is determined based on overall precedence of the sentences of one cluster over the others. 

\item \textit{Average position ordering (APO):}
The sentences $\{s_{p}, s_{q}, \ldots , s_{z}\}$ in any cluster $C_{i}$
%$p$, $q$, $\ldots$ , $z$ 
are each assigned a normalized score. For example, the normalized score of $s_{p}$ is computed as the ratio of the original position of the sentence and the total number of sentences in $d_{i}$ (here, $s_{p}$ belongs to document $d_{i}$). %Here, $d_{i}$ is the document, to which $S_{p}$ belongs. 
When ordering two clusters, the cluster that has the lower score obtained by averaging the normalized scores of all the sentences  %is assigned precedence over the other. 
is ranked higher than the others. 
\end{enumerate}

%Kaz-IJCAI-1: I think ``S2: ...'' looks better than ``3.2 S1: ...'' This is the same reason as ``S1: ...'' 
\subsection*{S2: Summary Sentence Generation}
\label{sec:summsentence}
\begin{figure}[t]
 \centering
 %\fbox{ \includegraphics[width=0.45\textwidth,height=2.5in, keepaspectratio=false]{word-graph-generation.png}}
 \fbox{\includegraphics[width=0.45\textwidth, height=2in]{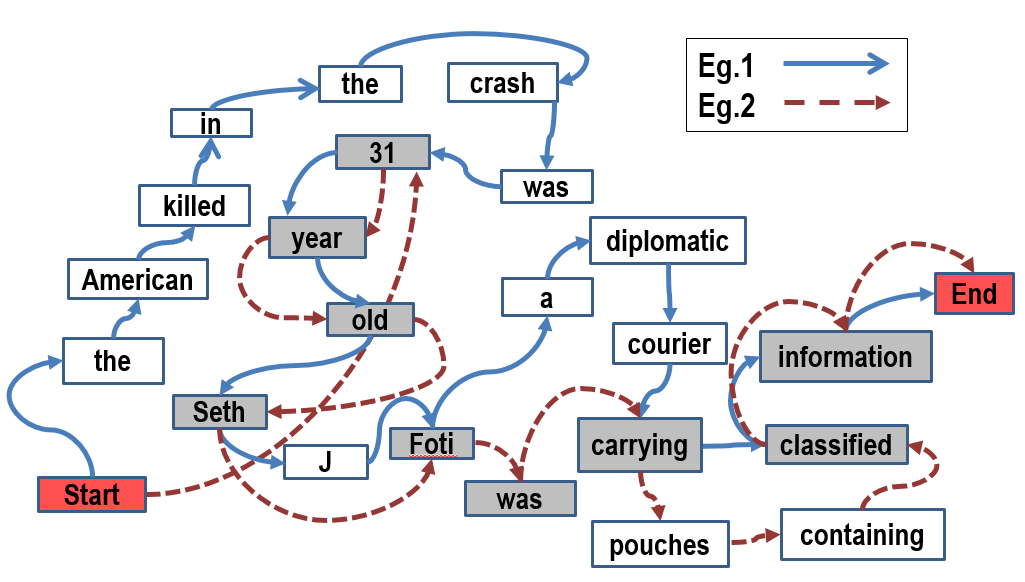}}
\caption{Word-graph generation from sentences.}
\label{fig:word-graph-generation}
\end{figure}

In order to generate a one-sentence representation from a cluster of redundant sentences, we use multi-sentence compression. %We generate $K$ shortest paths and determine the paths that should be retained to generated a well-formed informative summary. We set $K$ to 200.  
%
%\subsubsection{Shortest Paths Generation}\label{subsec:MltSntCmp}
We generate multiple sentences from a cluster using a word-graph~\cite{filippova2010multi}. Suppose that a cluster $C_{i}$ contains $j$ sentences, $S=\{s_{1}, s_{2}, \ldots , s_{j}\}$. 
A directed graph is created by adding sentences from $S$ to the graph in an iterative fashion. Each sentence is connected to dummy \textit{start} and \textit{end} nodes to mark the beginning and ending of the sentences. %The set of vertices or nodes are the words in the sentences, 
%Kaz-IJCAI-1: I think that you do not need ``set of'' here.
The vertices or nodes are the words along with the parts-of-speech (POS) tags. 
We connect adjacent words in the sentences with directed edges.
Once the first sentence is added, words from the following sentences are mapped onto a node in the graph provided that
they have exactly the same word form and the same POS tag.
%Each word in the sentence is represented as the combination of the word and  to prevent ungrammatical mappings.
The sequence of rules used for the word-graph construction is as follows:

\small{
\begin{itemize}
\setlength\itemsep{0em}
%\item Non-stopwords are added for which there are no candidates in the already existing graph, or in case an unambiguous mapping is possible.
\item Content words are added for which there are no candidates in the existing graph,
%\item Non-stopwords for which there are several possibilities or such words that occur more than once in the sentence.
\item Content words for which multiple mappings are possible or such words that occur more than once in the sentence,
\item Stopwords.  
\end{itemize}
}\normalsize
The context of the words are taken into consideration if multiple mappings are possible, and the word is mapped to that node that has the highest directed context.
We also add punctuations to the graph. %We do not lowercase the words if they are proper nouns. 
Figure~\ref{fig:word-graph-generation} shows a simple example of the word-graph generation technique.
We do not show POS and punctuations in the figure for clarity.
% Suppose the following two sentences for simplicity:
Consider the following two sentences as an illustration of our generation approach:

\small{
\begin{itemize}
\setlength\itemsep{0em}
\item[\textbf{Eg.1}] \textit{The American killed in the crash was 31-year-old Seth J. Foti, a diplomatic courier carrying classified information.}
\item[\textbf{Eg.2}] \textit{31-year-old Seth Foti was carrying pouches containing classified information.}
\end{itemize}}
\normalsize
As shown in the examples above, the two sentences contain similar information, but they are syntactically different. 
The solid directed arrows connect the nodes in Eg.1, whereas the dotted arrows join the nodes in Eg.2. 
%As can be seen, several shortest paths can be obtained in between the start and end nodes. 
We can obtain several shortest paths between the start and end nodes. 
%One such example has been highlighted in \textit{grey} in the figure. 
In Figure~\ref{fig:word-graph-generation}, we highlight one such path using gray rectangles. Several other paths are possible, for example:

\small{
\begin{itemize}
\setlength\itemsep{0em}
 \item[1:] \textit{31 year old Seth Foti was carrying pouches containing classified information.}
 \item[2:] \textit{31 year old Seth Foti a diplomatic courier carrying classified information.}
\end{itemize}}
\normalsize
%Our aim is to select the sentence that provides the maximum amount of information and is grammatically coherent among these set of several possible paths. 
The original input sentences from the cluster are also valid paths between the \textit{start} and \textit{end} nodes. To ensure pure abstractive summarization, we remove such paths that are same or very similar (cosine similarity $\ge$ 0.8) %(cosine similarity - 0.8 and above) 
to any of the original sentences in the cluster.
Similar to Filippova's word-graph construction, we set the minimum path length (in words) to eight to avoid incomplete sentences.
Finally, we retain a maximum of 200 randomly selected paths from each cluster to reduce computational overload of the ILP based approach. 
Our aim is to select the best path from all available paths.

\subsubsection{Best Path Selection} 
From 200 paths in each cluster, we choose at most one path that maximizes information content and linguistic quality together. 
%Let us denote each path in a cluster $C_{j}$ by $p^{C_{j}}_{i}$. Hence,
Let $p^{C_{j}}_{i}$ be each path in a cluster $C_{j}$, namely, 

\begin{center}{$\forall j \in \{1,....,m\}, C_{j} \ni \{p^{C_{j}}_{1}, p^{C_{j}}_{2},... ,p^{C_{j}}_{K}\}$},\end{center}
where the total number of shortest paths is equal to $K$=$\min[|C_{j}|, 200]$ where $|C_{j}|$ refers to the maximum number of paths that can be generated from a cluster. %In some cases, the number of shortest paths generated might be less than 200. 
We argue that the shortest paths that we select in the final summary should be informative as well as linguistically readable. Hence, we introduce two factors 
-- \textit{Informativeness} ($I(p^{C_{j}}_{i})$) and \textit{Linguistic quality} ($LQ(p^{C_{j}}_{i})$).

\begin{table*}[t]
\centering
\scriptsize
\caption{\footnotesize{
Comparison of ROUGE scores on the DUC 2004 and 2005 datasets: Baselines, state-of-the-arts, our proposed methods and abstractive summarization system using MSC. ``$\dagger$'' denotes the differences between [ILPSumm] and the baselines on the ROUGE scores are statistically significant for $p<0.05$.
%We compared ROUGE-2 and ROUGE-SU4 scores on the DUC2004 dataset. However, we compared ROUGE-L and ROUGE-SU4 scores on the DUC 2005 dataset as the other scores were not reported in previous work. 
We limit ROUGE evaluation to 665 bytes and 250 words for the 2004 and 2005 datasets, respectively. 
%``*'' denotes the differences between [ILPSumm] and CLASSY04/OCCAMSV/CLASSY11 systems on all ROUGE scores are statistically significant for $p<0.05$. \scriptsize{(Note that the lower the divergence scores (KL-Div and JS-Div), 
%the better the quality of summaries).} 
}}
\label{table:comprouge}
\begin{tabular}{l|cc||l|cc}
\hline
\multicolumn{3}{c||}{\textbf{\normalsize{DUC-2004}}} & \multicolumn{3}{c}{\textbf{\normalsize{DUC-2005}}}\\
\hline
 &  \textbf{ROUGE-2} & \textbf{ROUGE-SU4}&   &\textbf{ROUGE-L} & \textbf{ROUGE-SU4}\\
\hline
\textbf{Baselines}& &&\textbf{Baselines}&&\\
\hline
GreedyKL & 0.08658 & 0.13253 & Random&0.26395&0.09066 \\
FreqSum & 0.08218 & 0.12448 & Centroid&0.32562&0.11007\\
Centroid &  0.08139 & 0.12642 &LexRank&0.33179&0.12021\\
TsSum & 0.08068 & 0.12209&&&\\
LexRank &  0.07796 & 0.12484&&&\\
\hline
\textbf{State-of-the-arts}& &&\textbf{State-of-the-arts}&&\\
\hline
DPP    & {0.10079} & {0.14556} & DUC Best & 0.34764 & 0.10012\\
Submodular  & 0.09602 & {0.14227} & LSA & 0.26476 &0.10806\\
RegSum  & \textit{0.09712} & 0.13812 & NMF&0.28716& 0.11278\\
CLASSY04 & 0.09168 & 0.13250 &KM&0.29107& 0.10806\\
OCCAMSV & 0.09420 & 0.13105 &FGB&0.35018&0.12006\\
ICSISumm  & 0.09585 & 0.13314 &RDMS&0.35376&0.12297\\
CLASSY11 & 0.08912 & 0.12779&&&\\
\hline
%${MD^{Imp}_{LexRank}}$ + MO  & $0.38891^{\dagger}$ & $0.09612^{*,\dagger}$  & $\textit{0.34676}^{*,\dagger}$  & $0.13911^{\dagger}$ & $\textit{0.12032}^{\dagger}$  \\
\textbf{Abstractive Systems}\\
\hline
${MD^{Imp}_{LexRank}}$ + APO + MSC [Filippova, 2010] & 0.09612  &0.13911&${MD^{Imp}_{LexRank}}$ + APO + MSC&0.35589&0.12211\\
${MD^{Imp}_{LexRank}}$ + MO + ILP &0.09799&0.13884&${MD^{Imp}_{LexRank}}$ + MO + ILP&0.35281&0.12107\\
${MD^{Imp}_{LexRank}}$ + APO + ILP &0.10317&0.14218&${MD^{Imp}_{LexRank}}$ + APO + ILP&0.35342&0.12117\\
${MD^{Imp}_{CosSim}}$ + MO + ILP&0.09799&0.13884&${MD^{Imp}_{CosSim}}$ + MO + ILP&0.35661&0.12331\\
${MD^{Imp}_{CosSim}}$ + APO + ILP&0.10577&0.14215&${MD^{Imp}_{CosSim}}$ + APO + ILP&0.35577&0.12298\\
${MD^{Imp}_{DocsetSim}}$ + MO + ILP \textbf{[ILPSumm]}&$\textbf{0.11992}^{\dagger}$&$\textbf{0.14765}^{\dagger}$&${MD^{Imp}_{DocsetSim}}$ + MO + ILP \textbf{[ILPSumm]}&$\textbf{0.35772}^{\dagger}$&$\textbf{0.12411}^{\dagger}$\\
${MD^{Imp}_{DocsetSim}}$ + APO + ILP&0.11712&0.13567&${MD^{Imp}_{DocsetSim}}$ + APO + ILP&0.35679&0.12393\\
\hline
\hline
\end{tabular}%
\end{table*}

\noindent\textbf{Informativeness:} %In theory, 
In principle, we can use any existing method that computes the importance of a sentence to define \textit{Informativeness}. In our model, we use TextRank scores~\cite{mihalcea2004textrank} to generate an importance value of a sentence within a cluster. TextRank creates a graph of words from the sentences. The score of each node in the graph is calculated as shown in Equation~(\ref{lpr-textrank}): 

\small{
\begin{equation}
S(V_{i}) = (1-d) + d\times\sum_{V_{j}\in{}adj(V_{i})}\frac{w_{ji}}{\sum_{V_{k}\in adj(V_{i})}w_{jk}}S(V_{i}), 
\label{lpr-textrank}
\end{equation}}
\normalsize
where $V_{i}$ represents the words, $adj(V_{i})$ denotes the adjacent nodes of $V_{i}$ and $d$ is the damping factor set to 0.85. The computation converges to return final word importance scores. The informativeness score of a path ($I(p^{C_{j}}_{i})$) is obtained by adding the importance scores of the individual words in the path. 
%Possible candidate keyphrases are scored by summing up the importance scores of the words it contains.

\noindent\textbf{Linguistic Quality:} In order to compute \textit{Linguistic quality}, we use a language model. More specifically, we use a 3-gram (trigram) language model that assigns probabilities to sequence of words. Suppose that a path contains a sequence of $q$ words $\{w_{1},w_{2},...,w_{q}\}$. The score $LQ(p^{C_{j}}_{i})$ assigned to each path is defined as follows:

\small{
\begin{equation}
LQ(p^{C_{j}}_{i}) =\frac{1}{1 - LL(w_{1},w_{2},\ldots,w_{q})}, 
\label{eqn:ll2}
\end{equation}}
\normalsize
where $LL(w_{1},w_{2},...,w_{q})$ is defined as: 

\small{
\begin{equation}
LL(w_{1},w_{2},\ldots,w_{q}) = \frac{1}{L}\cdot\log_2 \prod_{t=3}^{q} P(w_{t}|w_{t-1}w_{t-2}). 
\label{eqn:ll}
\end{equation}}
\normalsize

As can be seen from Equation (\ref{eqn:ll}), we obtain the conditional probability of different sets of 3-grams in the sentence. The scores are combined and averaged by $L$, the number of conditional probabilities computed. The $LL(w_{1},w_{2},\ldots,w_{q})$ scores are negative; with higher magnitude implying lower readability. Therefore, in Equation~(\ref{eqn:ll2}), we take the reciprocal of the logarithmic value with smoothing to compute $LQ(p^{C_{j}}_{i})$. In our experiments, we used a 3-gram model that is trained on the English Gigaword corpus\footnote{\scriptsize{The model is available here: \url{http://www.keithv.com/software/giga/}. We used the VP 20K vocab version.}}.
\subsubsection{ILP Formulation}
To select the best paths from the clusters, we combine informativeness $I(p^{C_{j}}_{i})$ and linguistic quality $LQ(p^{C_{j}}_{i})$ in an optimization framework. We maximize the following objective function:

\small{
\begin{equation}
F(p^{C_{1}}_{1},\ldots , p^{C_{m}}_{K})=\sum_{j=1}^m \sum_{i=1}^{K}\frac{1}{T(p^{C_{j}}_{i})} \cdot I(p^{C_{j}}_{i}) \cdot LQ(p^{C_{j}}_{i}) \cdot p^{C_{j}}_{i} 
\label{eqn:obj}
\end{equation}}
\normalsize
%The factors of content and readability has been modeled into the objective function. 

Each $p^{C_{j}}_{i}$ represents a binary variable, that can take 0 or 1, depending on whether the path is selected in the final summary or not. 
In addition, $T(p^{C_{j}}_{i})$, the number of tokens in a path, is also taken into consideration and the term $\frac{1}{T(p^{C_{j}}_{i})}$ assigns more weight to shorter paths 
so that the system can favor shorter informative sentences. We introduce several constraints to solve the problem. 
First, we ensure that a maximum of one path is selected from each cluster using Equation~(\ref{eqn:maxone}). 

\small{
\begin{equation}
\forall j \in \{1,\ldots, m\}, \sum_{i=1}^{K} p^{C_{j}}_{i} \leq 1
\label{eqn:maxone}
\end{equation}}
\normalsize
%To prevent similar information being selected from different clusters, 
\noindent We introduce Equation~(\ref{eqn:sim}) so that we can prevent similar information (cosine similarity $\ge$ 0.5) from being selected from different clusters. 
In Figure~\ref{fig:word-graph-generation}, this constraint ensures that only one of the several possible paths mentioned in the example is included in the final summary as they contain redundant information.
%Kaz-IJCAI-1: It's better to write ``if sim() \ge 0.5'' in new line. 

\small{
\begin{eqnarray}
\forall j,j' \in [1,m], i,i' \in [1,K] \subset C_{j},C_{j'} \nonumber\\
p^{C_{j}}_{i} + p^{C_{j'}}_{i'} \leq 1 \text{ if } sim(p^{C_{j}}_{i},p^{C_{j'}}_{i'}) \geq 0.5. 
\label{eqn:sim}
\end{eqnarray}}

\normalsize
\section{Experimental Results}\label{Sec:ExpResults}
\subsection{Dataset and Evaluation Metrics}
We evaluated our approach on the DUC 2004 and 2005 datasets\footnote{\url{http://duc.nist.gov/data.html}} on multi-document summarization. 
We use ROUGE (Recall-Oriented Understudy of Gisting Evaluation)~\cite{lin2004rouge} for automatic evaluation of summaries (compared against human-written model summaries) as it has been proven effective in measuring qualities of summaries and correlates well to human judgments. %We also use automatic evaluation using information-theoretic measures by comparing the summaries against the input documents~\cite{louis2009automatically}. Further, in order to understand the overall quality of the abstractive summaries, we also perform a manual evaluation.
%\subsubsection{ROUGE-based Comparison: Baselines and State-of-the-arts}
%The summaries generated by the baselines and the state-of-the-art extractive summarizers on the DUC 2004 data were collected from Hong et al.,~\shortcite{hong2014repositary}. 
%We collected ROUGE scores on the DUC 2005 dataset from Zheng et al.~\cite{zheng2014multi}. 
\subsection{ROUGE Evaluation} \label{subsec:comp}
%We compare ROUGE scores obtained by our systems with those obtained by some baselines and the state-of-the-art extractive summarizers as reported by Hong
%et al.,~\shortcite{hong2014repositary} on the DUC 2004 dataset. 
We proposed three document importance measures and two different sentence ordering techniques as described in Section~\ref{Sec:Method}. Hence, we have six different systems in total. 
To the best of our knowledge, no publicly available abstractive summarizers have been used on the DUC dataset. 
Therefore, we compare our system to MSC~\cite{filippova2010multi} that generates a sentence from a collection of similar sentences using only syntactical information from the source sentences. In MSC, the input is a pre-defined cluster of similar sentences. Therefore, we compare our ILP based technique with MSC using the same set of input clusters obtained by our system. 
Table~\ref{table:comprouge} shows the following ROUGE scores for our evaluation: 

\small{
\begin{itemize}
\setlength\itemsep{0em}
 \item[] \textbf{ROUGE-2}: $N$-gram based ROUGE, where $N$=2, denotes the size of the sequence of words. %where $n$ is the length of the $n$-gram), \\
 \item[] \textbf{ROUGE-L} (longest common subsequence),
 \item[] \textbf{ROUGE-SU4}: skip-bigrams with unigrams.
\end{itemize}}

\normalsize
The summaries generated by the baselines and the state-of-the-art extractive summarizers on the DUC 2004 data were collected from ~\cite{hong2014repositary}. 
ROUGE-2 and ROUGE-SU4 scores have been found to be highly correlated with human judgments~\cite{nenkova2011automatic}.
Therefore, we computed ROUGE-2 and ROUGE-SU4 scores of the other systems on the DUC2004 summaries directly using ROUGE. However, the system-generated summaries (baselines and state-of-the-arts) were not available for the DUC 2005 dataset. Hence, we used ROUGE scores of the various systems as reported in ~\cite{zheng2014multi}. 

According to Table~\ref{table:comprouge}, all of the ROUGE 
scores obtained by our systems outperform all the baselines on both datasets. 
Hereafter, we refer to the best performing system as \textbf{ILPSumm}.
%We also did a paired t-test and found our best performing systems to have significant differences from the baselines on ROUGE-2 and ROUGE-L scores. 
We perform paired t-test and observe that ILPSumm shows statistical significance compared to all the baselines. %  and some state-of-the-art extractive systems (CLASSY04/OCCAMSV/CLASSY11) on all ROUGE scores ($p<0.05$).
The summarization method using ${MD^{Imp}_{DocsetSim}}$ measure as the most informative document and ranked by majority ordering (MO) outperforms all of the other techniques. 
The document that has the highest similarity to the total content captures the central idea of the documents. % collection. %This point onward, 
The clustering scheme that works best with MSC is ${MD^{Imp}_{LexRank}}$ + APO. ILPSumm also outperforms the MSC-based method, i.e., our approach can generate more informative summaries by globally maximizing content selection from multiple clusters of sentences. In summary, content selection of our proposed abstractive systems work at par with the best extractive systems.

\noindent \textbf{Discussion:} Our proposed system identifies the most important document, which is a general human strategy for summarization. The majority ordering strategy prioritizes 
clusters that contain sentences which should be mentioned earlier in a summary. %The ILP solution ensures that the most important content in each cluster is conveyed in the generated sentence. 
Other systems tackle redundancy as a final step; however, we integrate linguistic quality and informativeness to select the best sentences in the summary using our ILP based approach.

We performed the rest of our experiments only on the DUC 2004 dataset as it has been widely used for multi-document summarization. 
\subsection{Manual Evaluation}
We also determine readability of the generated summaries by obtaining ratings from human judges. Following Liu and Liu~\shortcite{liu2009extract}, we ask 10 evaluators to rate 10 sets of four summaries on two different factors -- \textit{informativeness} and \textit{linguistic quality}. The ratings range from 1 (lowest) to 5 (highest). All the evaluators have a good command of English and seven of them are native speakers. Evaluators were asked to rate the summaries based on informativeness (the amount of information conveyed) and linguistic quality (readability of the summary).  We randomized the sets of summaries to avoid any bias.  %Informativeness and linguistic quality measure the amount of information and readability of the summaries, respectively.
%Kaz-IJCAI-1: I think only ``LQ'' is better than ``LingQ'' as it's easier to understand. 
\begin{table}[t]
\centering
\footnotesize
\caption{\small{Manual evaluation by 10 evaluators on \textit{Informativeness} {(Inf)} and \textit{Linguistic Quality} {(LQ)} of summaries. Average Log-likelihood scores (Avg.LL) from parser are also shown.}}
\label{tab:humaneval}
\begin{tabulary}{0.52\textwidth}{L|CC|C}
\hline
\textbf{Type} & \textbf{Inf} & \textbf{LQ} & \textbf{Avg.LL}\\
\hline
Human written & 4.42  & 4.35 & -129.02\\
Extractive (DPP) & 3.90  & 3.81 & -142.70\\
Abstractive (MSC) & 3.78  & 2.83 &-210.02\\
Abstractive (ILPSumm) & 4.10  & 3.63 &-180.76\\
\hline
\end{tabulary}%
\end{table}

The four summaries provided to the evaluators are human-written summary (one summary collected randomly from four model-summaries per cluster), extractive summary (DPP), abstractive summary generated using MSC (MSC) and abstractive summary generated using our ILP based method (ILPSumm). 
We asked each evaluator to complete 10 such tasks, each containing four summaries as explained above. 
We normalize ratings of different evaluators to the same scale.
Table~\ref{tab:humaneval} shows the results obtained by manual evaluation. 
According to the judges, the linguistic quality of \textbf{ILPSumm} (3.63) is significantly better than that of \textbf{MSC} (2.83). Further, our summaries (ILPSumm) are more informative than DPP (3.90) and MSC (3.78). DPP is extractive in nature, hence linguistically, the sentences are generally more readable. To obtain a coarse estimate of grammaticality, we also compute the confidence scores of the summaries using the Stanford Dependency Parser~\cite{chen2014fast}. A language model assigns probabilities to sequence of words; in contrast, the confidence score of a parser assigns probabilities to grammatical relations. The values (the lower the magnitude, the better) are shown in the column \textit{Avg.LL}.
\textit{Avg.LL} obtained by ILPSumm (-180.76) is better than that obtained by MSC (-210.02), indicating that the language model based linguistic quality estimation helps generate more readable summaries than the MSC method.

Table~\ref{tab:sampleSumm} shows a comparison of summaries from the different systems using the DUC 2004 dataset. As can be seen, the linguistic quality of the abstractive summaries (ILPSumm) is acceptable, and the content is well-formed and informative. Our ILP framework can combine information from various sentences and present a fairly well-formed readable summary.

\begin{table}[t] % no need to specify "[ht]"
\centering
\scriptsize
%\caption{Sample summaries - comparing abstractive summaries with extractive summaries and human-written summaries}
\caption{\small{Example of DUC2004 abstractive summaries obtained using our approach (ILPSumm), extractive summaries (DPP) and and human-written summaries. Only a few initial sentences are displayed here.}
}
\label{tab:sampleSumm}
\begin{tabular}{p{0.45\textwidth}}%
%\newline%
\hrule
\textbf{Abstractive summary (ILPSumm):} Hun Sen's Cambodian People's Party won 64 of the 122 parliamentary seats in July.
Opposition ally Sam Rainsy charged that Hun Sen's party has rejected allegations of intimidation and fraud .
Hun Sen and Ranariddh are to form working groups this week to divide remaining government posts.
But a deal reached between Hun Sen and his chief rival, Prince Norodom Ranariddh's ally, Sam Rainsy.\\
\textbf{Extractive summary (DPP):} Ranariddh and Sam Rainsy have charged that Hun Sen's victory in the elections was achieved through widespread fraud.
Hun Sen said his current government would remain in power as long as the opposition refused to form a new one. 
Cambodian leader Hun Sen, who heads the CPP, has offered to share the legislature's top job with the royalist FUNCINPEC party of Prince Norodom Ranariddh in order to break the impasse.\\
\textbf{Human-written summary:} Cambodian prime minister Hun Sen rejects demands of 2 opposition parties for talks in Beijing after failing to win a 2/3 majority in recent elections.
Sihanouk refuses to host talks in Beijing.
Opposition parties ask the Asian Development Bank to stop loans to Hun Sen's government.
CCP defends Hun Sen to the US Senate.
FUNCINPEC refuses to share the presidency.
Hun Sen and Ranariddh eventually form a coalition at summit convened by Sihanouk.
%\hline
\hrule%
\hrule%
\textbf{Abstractive Summary (ILPSumm):} Lebanese Foreign Minister Kamal Kharrazi made the mediation offer Sunday, in a telephone conversation with his Syrian counterpart, Farouk al-Sharaa.
Egyptian President Hosni Mubarak met here Sunday with Syrian President Hafez Assad to show Lebanon's support for Syria and Turkey.
In a show of force on Friday, Turkish troops were deployed this week on the Turkish-Syrian border to eradicate Krudish rebel bases.\\
\textbf{Extractive Summary (DPP):} Egyptian President Hosni Mubarak met here Sunday with Syrian President Hafez Assad to try to defuse growing tension between Syria and Turkey. The talks in Damascus came as Turkey has massed forces near the border with Syria after threatening to eradicate Kurdish rebel bases in the neighboring country. Egypt already has launched a mediation effort to try to prevent a military confrontation over Turkish allegations that Syria is harboring Turkish Kurdish rebels.\\
\textbf{Human-written Summary:} Tensions between Syria and Turkey increased as Turkey sent 10,000 troops to its border with Syria.
The dispute comes amid accusations by Turkey that Syria helping Kurdish rebels based in Syria.
Kurdish rebels have been conducting cross border raids into Turkey in an effort to gain Kurdish autonomy in the region.
\hrule%
%\hline
\end{tabular}
\end{table}

\noindent{\textbf{Error Analysis:}} There is still room for improvement in the linguistic quality of the generated
summaries. We analyzed the summaries that were given lower ratings than the other options on the basis of linguistic quality. Consider the following sentence in a system generated summary, which received low scores from human judges:
\justify{
\textit{\small{The U.N. imposed sanctions since 1992 for its refusal to hand over the two Libyans wanted in the 1988 bombing that killed 270 people killed.}}}

\noindent As can be seen, the phrase ``killed 270 people killed '' is not coherent. The language model fails to identify such cases as the 3-gram sequences of \textit{killed 270 people} and \textit{270 people killed} are both grammatically coherent. In addition to a language model, we can also use a dependency parser to assign lower weights to paths that have redundant dependencies on the same nodes.  
Consider the following example:
\justify{\textit{\small{The deal that will make Hun Sen prime minister and Ranariddh agreed to a government formed}.}}\\

\noindent The last phrase -- ``a government formed'', is grammatically incoherent in the context of the sentence. Linguistically correct modifications could be -- \textit{a government being formed} or \textit{a government formation}. In future work, we plan to address such issues of grammaticality using dependency parses of sentences rather than just adjacency relations when building the word-graph. 

\section{Conclusions and Future Work}\label{Sec:Conc}
%In this work, 
We have proposed an approach to generate abstractive summaries from %a collection of text documents. 
a document collection. 
%We take a step forward towards multi-document abstractive summarization that 
We capture the redundant information using a simple yet effective clustering technique. We proposed a novel ILP based technique to select the best shortest paths in a word-graph to maximize information content and linguistic quality of a summary. 
Experimental results on the DUC 2004 and 2005 datasets show that %the effectiveness of 
our proposed approach outperforms all the baselines and the state-of-the-art extractive summarizers. %extractive summarization systems. 
Based on human judgments, our abstractive summaries are linguistically preferable than the baseline abstractive summarization technique. 
%Moreover, the system generated summaries are more informative than those obtained by the extractive models.
In future work, we plan to use paraphrasing %rewriting 
techniques to further enhance quality of the generated summaries. We also plan to address phrase level redundancies to improve coherence.%We also plan to develop better measures for automatic evaluation of abstractive summaries. Furthermore, 
%We can also add more constraints to our ILP problem to limit the length of generated summaries.
%\vspace{-4mm}
%\newpage
\section*{Acknowledgments}
This material is based upon work supported by the National Science Foundation under Grant No. 0845487.

\small{
\bibliographystyle{named}
\bibliography{ijcai16} } % sigproc.bib is the name of the Bibliography in this case

\begin{thebibliography}{}

\bibitem[\protect\citeauthoryear{Barzilay and
  Mckeown}{2005}]{Barzilay05sentencefusion}
Regina Barzilay and Kathleen~R. Mckeown.
\newblock {Sentence Fusion for Multidocument News Summarization}.
\newblock {\em Computational Linguistics}, 31(3):297--328, 2005.

\bibitem[\protect\citeauthoryear{Berg-Kirkpatrick \bgroup \em et al.\egroup
  }{2011}]{Berg-Kirkpatrick:2011:JLE:2002472.2002534}
Taylor Berg-Kirkpatrick, Dan Gillick, and Dan Klein.
\newblock {Jointly Learning to Extract and Compress}.
\newblock In {\em Proc. of the 49th Annual Meeting of the Association for
  Computational Linguistics (ACL-HLT 2011)}, pages 481--490, 2011.

\bibitem[\protect\citeauthoryear{Chen and Manning}{2014}]{chen2014fast}
Danqi Chen and Christopher~D Manning.
\newblock {A Fast and Accurate Dependency Parser using Neural Networks}.
\newblock In {\em Proc. of the 2014 Conference on Empirical Methods in Natural
  Language Processing (EMNLP)}, pages 740--750, 2014.

\bibitem[\protect\citeauthoryear{Erkan and Radev}{2004}]{erkan2004lexrank}
G{\"u}nes Erkan and Dragomir~R Radev.
\newblock {LexRank: Graph-based Lexical Centrality as Salience in Text
  Summarization}.
\newblock {\em Journal of Artificial Intelligence Research (JAIR)},
  22:457--479, 2004.

\bibitem[\protect\citeauthoryear{Filippova}{2010}]{filippova2010multi}
Katja Filippova.
\newblock {Multi-Sentence Compression: Finding Shortest Paths in Word Graphs}.
\newblock In {\em Proc. of the 23rd International Conference on Computational
  Linguistics (Coling 2010)}, pages 322--330, 2010.

\bibitem[\protect\citeauthoryear{Ganesan \bgroup \em et al.\egroup
  }{2010}]{ganesan2010opinosis}
Kavita Ganesan, ChengXiang Zhai, and Jiawei Han.
\newblock {Opinosis: A Graph-Based Approach to Abstractive Summarization of
  Highly Redundant Opinions}.
\newblock In {\em Proc. of the 23rd International Conference on Computational
  Linguistics (Coling 2010)}, pages 340--348, 2010.

\bibitem[\protect\citeauthoryear{Genest and Lapalme}{2010}]{genest2010text}
Pierre-Etienne Genest and Guy Lapalme.
\newblock {Text Generation for Abstractive Summarization}.
\newblock In {\em Proc. of the 3rd Text Analysis Conference (TAC 2010)}, 2010.

\bibitem[\protect\citeauthoryear{Hong \bgroup \em et al.\egroup
  }{2014}]{hong2014repositary}
Kai Hong, John~M Conroy, Benoit Favre, Alex Kulesza, Hui Lin, and Ani Nenkova.
\newblock {A Repository of State of the Art and Competitive Baseline Summaries
  for Generic News Summarization}.
\newblock In {\em Proc. of the 9th International Conference on Language
  Resources and Evaluation (LREC'14)}, pages 1608--1616, 2014.

\bibitem[\protect\citeauthoryear{Knight and
  Marcu}{2002}]{knight2002summarization}
Kevin Knight and Daniel Marcu.
\newblock {Summarization beyond Sentence Extraction: A Probabilistic Approach
  to Sentence Compression}.
\newblock {\em Artificial Intelligence}, 139(1):91--107, 2002.

\bibitem[\protect\citeauthoryear{Lin}{2004}]{lin2004rouge}
Chin-Yew Lin.
\newblock {ROUGE: A Package for Automatic Evaluation of Summaries}.
\newblock In {\em Proc. of the ACL-04 Workshop on Text Summarization Branches
  Out}, pages 74--81, 2004.

\bibitem[\protect\citeauthoryear{Liu and Liu}{2009}]{liu2009extract}
Fei Liu and Yang Liu.
\newblock {From Extractive to Abstractive Meeting Summaries: Can it be done by
  Sentence Compression?}
\newblock In {\em Proc. of the 47th Annual Meeting of the Association for
  Computational Linguistics (ACL-IJCNLP 2009)}, pages 261--264, 2009.

\bibitem[\protect\citeauthoryear{Mani and Maybury}{1999}]{mani1999advances}
Inderjeet Mani and Mark~T Maybury.
\newblock {\em {Advances in Automatic Text Summarization}}.
\newblock MIT Press, 1999.

\bibitem[\protect\citeauthoryear{Mehdad \bgroup \em et al.\egroup
  }{2013}]{mehdad2013abstractive}
Yashar Mehdad, Giuseppe Carenini, Frank~W Tompa, and Raymond~T NG.
\newblock {Abstractive Meeting Summarization with Entailment and Fusion}.
\newblock In {\em Proc. of the 14th European Workshop on Natural Language
  Generation}, pages 136--146, 2013.

\bibitem[\protect\citeauthoryear{Mihalcea and
  Tarau}{2004}]{mihalcea2004textrank}
Rada Mihalcea and Paul Tarau.
\newblock {TextRank: Bringing Order into Texts}.
\newblock In {\em Proc. of the Conference on Empirical Methods in Natural
  Language Processing (EMNLP 2004)}, pages 404--411, 2004.

\bibitem[\protect\citeauthoryear{Nenkova and
  McKeown}{2011}]{nenkova2011automatic}
Ani Nenkova and Kathleen McKeown.
\newblock Automatic summarization.
\newblock {\em Foundations and Trends in Information Retrieval},
  5(2-3):103--233, 2011.

\bibitem[\protect\citeauthoryear{Oya \bgroup \em et al.\egroup
  }{2014}]{templateSumm2014}
Tatsuro Oya, Yashar Mehdad, Giuseppe Carenini, and Raymond Ng.
\newblock {A Template-based Abstractive Meeting Summarization: Leveraging
  Summary and Source Text Relationships}.
\newblock In {\em Proc. of the 8th International Natural Language Generation
  Conference (INLG 2014)}, pages 45--53, 2014.

\bibitem[\protect\citeauthoryear{Wan}{2008}]{wan2008exploration}
Xiaojun Wan.
\newblock {An Exploration of Document Impact on Graph-Based Multi-Document
  Summarization}.
\newblock In {\em Proc. of the 2008 Conference on Empirical Methods in Natural
  Language Processing (EMNLP 2008)}, pages 755--762, 2008.

\bibitem[\protect\citeauthoryear{Wang and Cardie}{2013}]{wang2013domain}
Lu~Wang and Claire Cardie.
\newblock {Domain-Independent Abstract Generation for Focused Meeting
  Summarization}.
\newblock In {\em Proc. of the the 51st Annual Meeting of the Association for
  Computational Linguistics (ACL 2013)}, pages 1395--1405, 2013.

\bibitem[\protect\citeauthoryear{Zajic \bgroup \em et al.\egroup
  }{2007}]{zajic2007multi}
David Zajic, Bonnie~J Dorr, Jimmy Lin, and Richard Schwartz.
\newblock {Multi-Candidate Reduction: Sentence Compression as a Tool for
  Document Summarization Tasks}.
\newblock {\em Information Processing \& Management}, 43(6):1549--1570, 2007.

\bibitem[\protect\citeauthoryear{Zheng \bgroup \em et al.\egroup
  }{2014}]{zheng2014multi}
Hai-Tao Zheng, Shu-Qin Gong, Hao Chen, Yong Jiang, and Shu-Tao Xia.
\newblock Multi-document summarization based on sentence clustering.
\newblock In {\em Neural Information Processing}, pages 429--436. Springer,
  2014.

\end{thebibliography}
\end{document}